\documentclass[letterpaper]{article}
\usepackage{aaai}
\usepackage{times}
\usepackage{helvet}
\usepackage{courier}
\usepackage{fancyhdr}
\newcommand{\code}[1]{\textsf{#1}}

\newcommand{\tuple}[1]{{\langle #1 \rangle}}
\usepackage{comment}
\usepackage{graphicx}
\usepackage{listings}
\usepackage{multirow}

\title{HATP: An HTN Planner for Robotics}
\author{Rapha\"{e}l Lallement$^{1,2}$ \and Lavindra de Silva$^{1}$ \and Rachid Alami$^1$ \\\\ $^1$CNRS, LAAS, \\ 7 avenue du Colonel Roche, \\ F-31400 Toulouse, France \\\\ $^2$Univ de Toulouse, INSA, LAAS, \\ F-31400 Toulouse, France}

\begin{document}
\maketitle

\section{Abstract}
Hierarchical Task Network (HTN) planning is a popular approach that cuts 
down on the classical planning search space by relying on a given 
hierarchical library of domain control knowledge. This provides 
an intuitive methodology for specifying high-level instructions on how 
robots and agents should perform tasks, while also giving the planner enough 
flexibility to choose the lower-level steps and their ordering. In this
paper we present the HATP (Hierarchical Agent-based Task Planner)
planning framework which extends the 
traditional HTN planning domain representation and semantics by making them more 
suitable for roboticists, and treating agents as ``first class'' entities
in the language. The former is achieved by allowing ``social rules'' to be
defined which specify what behaviour is acceptable/unacceptable by 
the agents/robots in the domain, and interleaving planning with geometric 
reasoning in order to validate online--with 
respect to a detailed geometric 3D world--the human/robot actions currently 
being pursued by HATP.\footnote{This work has been conducted within the EU ARCAS
project (http://www.arcas-project.eu/) funded by the E.C. Division FP7-IST under
Contract ICT 287617. We thank the anonymous reviewers for their feedback. The second author has now
moved to The University of Nottingham, Nottingham, UK.}

\section{Introduction}

Real-world robotics domains and problems offer natural testbeds for HTN
(Hierarchical Task Network) planning. The intuitive hierarchical 
representation used by such planners allows the often available expert 
knowledge about a domain to be included with relative ease to guide the 
search process. This guidance might be abstract steps detailing
how a task, such as cleaning a table full of different types 
of objects, should be performed by the robot, with sufficient flexibility 
over the more detailed steps and states---e.g. the final locations of 
objects on the shelf. In practice, the inclusion
of such search control knowledge makes HTN planning faster than classical
planning, which is particularly important when dealing with robots as they need to
be responsive to environmental changes involving other robots,
and more importantly, humans. 

In this paper we describe the HATP (Hierarchical Agent-based Task Planner)
HTN planner and show how it is particularly suited for use in robotics. 
HATP is based on SHOP \cite{bib:Nau1999}, but unlike this planner and 
other HTN planners such as Nonlin \cite{Nonlin76},
SHOP2 \cite{bib:Shop2-2003} and UMCP \cite{bib:Umcp94}, HATP offers
a user-friendly domain representation language inspired by popular programming
languages, making it easier for roboticists, and indeed computer scientists 
alike, to become quickly acquainted with the syntax and semantics. We 
give insights into a formal mapping from this HATP language into an equivalent 
classical representation, but leave the detailed treatment for a separate paper.
 
An important feature of HATP is that it treats agents as ``first-class 
entities'' 
in the domain representation language. It can therefore distinguish between
the different agents in the domain as well as between agents and the other 
entities such as tables and chairs. This facilitates a post-processing step in HATP that splits the final 
solution (sequence of actions) into multiple synchronised solution streams, one per agent,
so that the streams may be executed in parallel by the respective agents by synchronising when necessary.

The planning algorithm of HATP has also been 
extended in various ways. First, it incorporates a simple mechanism to take 
into account the (user-defined) cost of executing actions, so that instead 
of returning the first arbitrary solution found, it keeps searching until an 
optimal (least-cost) one is found.\footnote{The notion of optimality here is
``local'': HATP finds an optimal solution only from the set of 
HATP solutions obtained using the given methods.} Second, HATP has been extended to be 
more suitable for Human-Robot Interaction (HRI); in particular, ``social 
rules'' can be included by the user to define what the acceptable (and
unacceptable) behaviours of the agents are. Two examples are: what
sequences of steps should be avoided in final solutions, and a limit 
on the amount of time a person should spend waiting (and doing nothing). 
The rules are then used to 
filter out the primitive solutions found that do not meet the constraints.

Finally, there is much ongoing work on interleaving HATP with geometric 
planning algorithms, so as to validate online the actions being pursued 
by HATP, by consulting its geometric counterpart. This results in 
motion planning being performed by the geometric planner to check if 
the HATP action being planned is actually feasible in 
the real world, modelled in great detail via the 
Move3D \cite{Simeon_2001_Move3d} simulation environment. This integration
takes an important step towards interfacing HATP's
AI planning algorithms and techniques with the planning algorithms
and techniques more commonly used by roboticists. In this paper we 
summarise all of these extensions to HATP, and explicate how they 
make HATP particularly suited for the Robotics community.

\section{HTN Planning}

While classical planners such as 
STRIPS focus on achieving some goal state, Hierarchical Task Network (HTN) 
planners focus on solving \textit{abstract tasks}. 
We have found HTN planning to be
particularly useful for robotics applications, as it allows---the often
available---instructions from the domain expert to be included in the domain
as an intuitive hierarchy.
This helps guide the search, making it faster in general than classical
planning approaches, and thereby also more practical for real robots that
need to be responsive to environmental changes. 

The Hierarchical Agent-based Task Planner (HATP) is based on the 
popular ``totally-ordered'' HTN planning approach, which unlike 
``partially-ordered'' HTN planning allows calls to external functions---a 
necessity in our work. This is also highlighted as a feature in the SHOP 
\cite{bib:Nau1999} planner, on which HATP is based. The rest of this section
focusses on totally-ordered HTN planning.  


We define an HTN \textit{planning problem} as the 3-tuple $\tuple{d, s_0, 
\mathcal{D}}$, where $d$, the ``goal'' to achieve, is a sequence of primitive 
or abstract tasks, $s_0$ is the initial state, and $\mathcal{D}$ is an HTN
\textit{planning domain}. An operator is as in classical planning, and
actions are ground instances of operators. We generally use the terms
operator and action interchangeably in this paper.
An HTN planning domain is the pair $\mathcal{D} = \tuple{\mathcal{A}, \mathcal{M}}$ where 
$\mathcal{A}$ is a finite set of operators, and $\mathcal{M}$ is a finite 
set of HTN \textit{methods}.
A method is a 4-tuple consisting of: the name of the method, the abstract 
task that it needs to solve, a precondition specifying when the method is 
applicable, and a \textit{body} realising the ``decomposition'' of the
task associated with the method into more specific subtasks. Specifically,
the method-body is a sequence of primitive and/or abstract tasks.

The HTN planning process works by selecting applicable methods from $\mathcal{M}$ 
and applying them to abstract tasks in $d$ in a depth-first manner. In each 
iteration, this will typically result in $d$ becoming a ``more primitive'' 
sequence 
of tasks. The process continues until $d$ has only primitive tasks left, which
map to action names.
At any stage during planning if no applicable method can be found
for an abstract task, the planner essentially ``backtracks'' and tries an
alternative method for an abstract task refined earlier. 

In more detail, the main steps of the HTN planning process are the following:
in each iteration all ground instances are found of the methods available 
to decompose a chosen task from task network $d$; one such method instance
is chosen arbitrarily that is applicable (whose precondition holds) in the 
current state of the world; and the instance is applied to $d$ 
by basically replacing the chosen task with the subtasks in the 
method's body. The planner backtracks to choose an alternative method 
instance to one that was previously applied to $d$ only if that method instance
did not eventually
allow a complete (and successful) decomposition of the top-level
goal task(s).

\section{Features of HATP}
In this section we present our own encoding in HATP of the Dock-Worker Robots 
domain~\cite{bib:automated_planning}. In this domain,
there is a robot (R1) that can move and carry containers, and two crane-agents
(K1 and K2) that can lift and put down containers. Furthermore, there are two 
locations (L1 and L2), each containing two piles (P11 and P12 at L1, 
and P21 and P22 at L2) that can hold containers. The goal is to 
place the two containers C1 and C2 on piles P21 and P22, respectively.

\begin{figure}[h!t]
\centering
\includegraphics[width=\columnwidth]{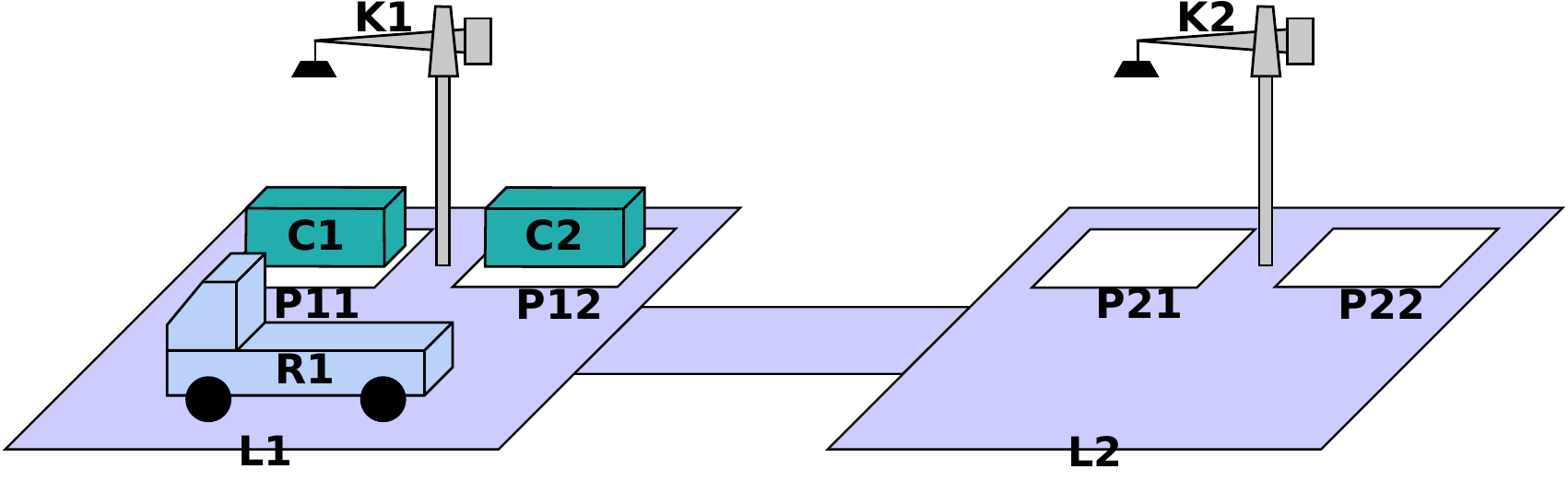}
\caption{A planning problem in the Dock-Worker Robot domain: robot R1 has 
to carry the container C1 from P11 to P21, and the container C2 from P12 to P22.}
\label{fig:dwr-problem}
\end{figure}

\subsection{World state representation}
In addition to having the standard advantages of total-order HTN planning, HATP
also provides an intuitive object-oriented-like syntax for
representing and manipulating the world state. This allows roboticists and
computer scientists alike to quickly get acquainted with the syntax and 
start developing HATP domains. 

The HATP world specification is defined
as a collection of entities, which represent the agent types and object
types in the world. This distinction between agents and other objects is
important. Agents are treated as first class entities in the language of
HATP; moreover, different types of agents may be defined by simply
instantiating the default Agent entity. This distinction also facilitates 
a post-processing step in HATP, which splits the final solution into 
separate sub-solutions to be executed by the respective agents. 

An entity has a set of attributes, where an attribute
either represents a data value, or a relation between the entity and other 
entities. For example, a robot-agent may have an attribute \code{carry} of type 
\code{Container} indicating that the robot can carry objects of type \code{Container}.
HATP supports some of the standard data types found in programming 
languages, such as integers and strings, and also allows defining sets of 
objects, which are manipulated using the standard set operations. An 
example of an HATP world specification is shown in listing~\ref{lst:entities}.\newline  \newline \newline

\begin{lstlisting}[
caption={HATP entities for the Dock-Worker Robot domain in figure~\ref{fig:dwr-problem}. There are five entity types: Agent (default entity), Crane, Location, Pile and Container. The initial value assigned to a \code{static} attribute cannot change during planning, whereas a \code{dynamic} attribute can be assigned different values over the course of planning. An attribute classified as an \code{atom} can only have one value, whereas one classified as a \code{set} can have a set of values. The \code{type} of an attribute can be any of the primitive types allowed as well as an entity.},
label={lst:entities}]
define entityType Crane, Location, Pile, Container;

define entityAttributes Agent {
  //An agent can be of type Robot or Crane 
  static atom string type;

  //For cranes
  static atom Location attached;

  //For robots
  dynamic atom Location at;
  dynamic atom Container carry;
  dynamic atom bool loading;
}

define entityAttributes Location {
  static set Location adjacent;
  dynamic atom bool occupied;
}

define entityAttributes Pile {
  static atom Location attached;
  dynamic set Container contains;
  dynamic atom Container top;
}

define entityAttributes Container {
  dynamic atom Location in;
  dynamic atom Container on;
}

\end{lstlisting}

The HATP initial world state is then an instantiation of the defined entities,
along with value assignments to their attributes. 
An example of an HATP initial world state is shown in listing~\ref{lst:state}.
Notice that attributes of entities generally map to predicate symbols in 
standard ``classical'' initial states, and the 
entities and values to the
parameters of the predicate. For example, \code{K1.attached = L1} could map to 
predicate \code{attached(K1,L1)} in a classical initial state, \code{R1.loading = false}
to \code{$\neg$loading(R1)}, and \code{R1.carry = NULL} could be represented in the
classical initial state by not
including any positive literal in it that has predicate symbol 
\code{carry}, with \code{R1} as its first parameter. 

\begin{lstlisting}[
caption={An HATP initial state for the Dock-Worker Robot domain. After instantiating the entity types, their attributes are assigned initial values. Note that symbol ``$<<=$'' is used to add the element on its RHS to the set on its LHS.},
label={lst:state}]
R1, K1, K2 = new Agent;
L1, L2 = new Location; 
P11, P12, P21, P22 = new Pile;
C1, C2 = new Container; 

R1.type = ``ROBOT'';
R1.at = L1;
R1.carry = NULL;
R1.loading = false;

K1.type = ``CRANE'';
K1.attached = L1;
K2.type = ``CRANE'';
K2.attached = L2;

L1.adjacent <<= L2;
L1.occupied = true;
L2.adjacent <<= L1;
L2.occupied = false;

P11.attached = L1;
P11.contains <<= C1;
P11.top = C1;
P12.attached = L1;
P12.contains <<= C2;
P12.top = C2;
P21.attached = L2;
P22.attached = L2;

C1.in = L1;
C1.on = NULL;
C2.in = L1;
C2.on = NULL;
\end{lstlisting}

\subsection{Domain representation}
As in standard HTN planning, an HATP domain consists of a set of methods and
a set of operators. These are written similarly to traditional HTN domains with 
the exception where the HATP language offers some user-friendly constructs 
for defining preconditions of methods and operators, 
bodies of methods and effects of operators. 
In particular, variables are defined in HATP methods, and their bindings
controlled, via the following constructs; examples of their use can be
found in listing~\ref{lst:method}.

\begin{itemize}
\item \code{SELECT} binds the given variable in the usual way. 
In essence, the construct amounts to a ``backtrack point'' that allows all
values of the associated variable---and thereby all ground instances of the 
method---to be considered.
\item \code{SELECTORDERED} binds the variable in some given order, governed
by a user-supplied ordering relation. Moreover, the variable can be bound 
in ascending or descending order with respect to the relation.  
\item \code{SELECTONCE} binds the variable only once---the remaining bindings
are disregarded. This offers a reduction in the branching factor at the 
expense of completeness, as some of the ignored bindings may also yield
HATP solutions.
\end{itemize}

While the last construct may result in the loss of HATP solutions, this 
heuristic is useful in domains where if a solution pursued
by taking one binding of the variable---and applying the resulting ground 
instance of the HATP method---turns out to not work, then no other binding 
for that variable will work either. For example, imagine a slightly different
Dock-Worker Robot domain/problem that has multiple robots, and where taking 
the shortest path during navigation is not important. This means that
if one robot cannot navigate from one location to another, then none of the
others will be able to either. Therefore, there is no need to consider
all possible robot-agent bindings as done in listing~\ref{lst:method}:
a single binding will be sufficient.
 
\begin{lstlisting}[
caption={Part of an HATP method to move a container from a source pile to a 
target pile in a different location. Note that \code{distance} is a 
user-supplied
ordering relation; ``$<$'' means that the variable bindings should be in descending 
order; and ``$N_1:T>N_2;$'' means that task $T$ (labelled $N_1$) must precede the 
task labelled $N_2$.}, 
label={lst:method}]
method Transport(Container C, Pile Target) {
  // do nothing if container is in target pile 
  empty{C.in == Target;};
  {
    preconditions {
      // container not already in target location 
      EXIST(Pile Source2, {C.in == Source2;}, 
	    {Source2.attached != Target.attached;});
    };
    subtasks {
      S = SELECT(Pile, {C.in == S;});
      R = SELECTORDERED(Agent, {R.type == ``ROBOT'';}, 
	  distance(R.at, S.attached), <);
      K1 = SELECT(Agent,
	   {K1.type == ``CRANE''; K1.at == S.attached;});
      K2 = SELECT(Agent, 
	   {K2.type == ``CRANE''; K2.at == Target.attached;});
      1: GetReady(R, C, S);
      2: LoadRobot(K1, R, C)>1;
      3: NavFromTo(R, S.attached, Target.attached)>2;
      4: UnloadRobot(K2, R, C)>3;
      5: Put(K2, C, Target)>4;
    };
    ...
}
\end{lstlisting}

Observe from listing~\ref{lst:method} that, as expected, the subtasks 
within the method's body are totally 
ordered. HATP, however, also allows partially ordering subtasks; this 
is achieved by
not specifying ordering constraints between some (or all) of the tasks in
the method's body. For example, removing constraint ``$>2$''
from the method in listing~\ref{lst:method} would then not require that the
task with label $3$ occur after the one with label $2$. Note that such partial 
ordering of tasks is merely a convenience: it is an alternative to supplying
multiple totally-ordered methods corresponding to every possible linearisation 
of the partially ordered subtasks. This is exactly what happens during 
planning: the set of 
partially ordered subtasks in a method's body is handled by taking all possible 
linearisations of the set, essentially creating additional HATP method options 
to consider for the parent task's decomposition.
Since partially ordering subtasks results in an exponential increase in the number
of method options, it should be used with appropriate care. Introducing ``true''
partially-ordered planning into HATP is left as future work: the algorithms are
not obvious as we want to have the ability to use evaluable predicates in
preconditions, for which maintaining the complete state of the world at each step of the
planning process is the obvious solution \cite{bib:Nau1999}.

Some other useful constructs supported by HATP are \code{EXIST}, 
\code{IF}, and \code{FORALL}. As in other HTN planners such as SHOP,
construct \code{EXIST} is used only in preconditions of methods and 
operators; \code{IF} only in the effects of operators; and \code{FORALL} 
in both preconditions of methods and operators, as well as in the effects 
of operators. 
Examples of how these constructs may be used are shown in 
listings \ref{lst:method} and \ref{lst:action}.

\begin{lstlisting}[
caption={An HATP operator. The expression ``$A >> B.attr$'' holds if element $A$ is in the set $B.attr$, and the expression's negation is specified using ``$A !>> B.attr$''. Expression ``$B.attr <<= A$'' adds element $A$ to set $B.attr$, and ``$B.attr =>> A$'' removes $A$ from $B.attr$.},
label={lst:action}]
action Move(
  Agent R, Location From, Location To, Location FinalDest) {
  preconditions {
    R.type == ``ROBOT'';
    To >> From.adjacent;
    R.at == From;
    To.occupied == false;
  };
  effects {
    R.at = To;
    From.occupied = false;
    To.occupied = true;
    R.path <<= To; 
    IF{From !>> R.path;}{R.path <<= From;} 
    IF{To.isForbiddenBy == R;}{To.isForbiddenBy = NULL;}
    IF{To == FinalDest;}{
      FORALL(Location LocP,
        {LocP >> R.path;},{R.path =>> LocP;});
    }
  };
  cost{costToMove(From, To)};
}
\end{lstlisting}

\subsection{Plan production}
HATP is able to find the least-cost primitive solution that solves the goal task(s)
at hand, as done for example in \cite{bib:Shop2-2003}. To this end,
HATP keeps track of the least-costly plan computed so far, as well as the
total cost of the current partial plan being pursued, and then avoids adding 
any action to it that will definitely lead to a costlier partial plan.
Indeed, in the worst case this requires looking through all HATP solutions
for the given goal task(s). Moreover, since the HATP search space is governed
by the methods supplied, there may be other low-cost solutions (corresponding to
methods not supplied) that HATP does not take into account. 
%
%

The cost of the partial plan is computed via ``cost functions''. A cost function
is a user-supplied C++ function that is linked to an HATP operator as shown at 
the bottom of listing~\ref{lst:action}. The function can perform any arbitrary 
calculation to estimate the cost of executing the 
action; however, for efficiency reasons the function should terminate quickly.
An example of such a function is one that computes the cost of 
executing an action to send data from one robot to another. This might involve checking
how much data needs to be sent and thereby how much time it would take
to do the transfer. 

By using cost functions associated with the sequence of
primitive actions pursued so far, HATP determines the total cost of the 
sequence, and avoids pursuing it further if by adding the next action 
the total cost would exceed 
the cost of the lowest-cost solution found so far.

Once HATP finds a solution---a sequence of primitive actions---it then splits
the solution into multiple ``streams'', one per agent in the 
domain, and adds causal links between streams for 
synchronisation \cite{bib:belief}.
To determine which actions in the final solution belong to which agents, the HATP
language reserves the first variable of every operator's name: it must always bind
to the name of the agent responsible for eventually executing the operator. 
The second and subsequent variables of an operator's name may also be used as
placeholders for agent names if necessary. Such an operator would then be a 
``joint operator'': one that needs to be executed in parallel by all the 
robots/agents that it refers to.

\begin{figure*}[ht]
\centering
\includegraphics[width=\textwidth]{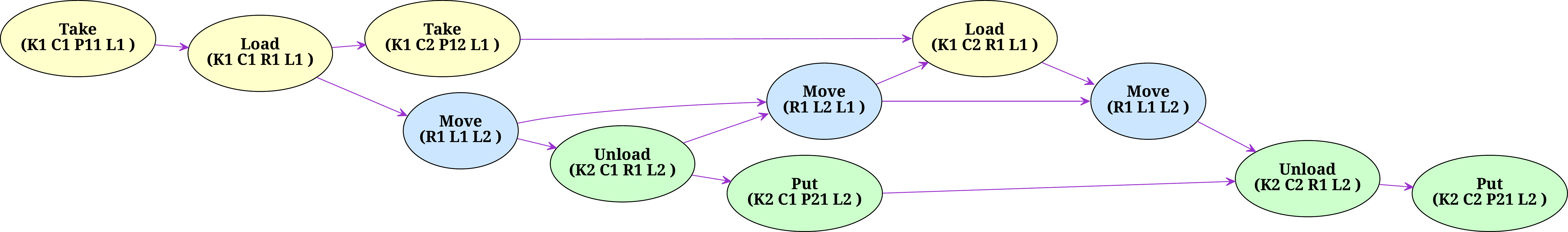}
\caption{The HATP solution for the DWR problem. There are three streams 
corresponding to the actions belonging to the three agents, ordered using 
causal links (shown as arrows). The yellow stream represents the actions 
of the first crane, the green the actions of the other crane, and the blue 
the actions of the robots. }
\label{fig:dwr-planstreams}
\end{figure*}

Once the different streams are separated, they may then be executed. The stream 
(if any) belonging to the agent that formulated the plan may be executed by the
agent directly, whereas
actions in other streams need to be delegated to their respective agents, and
the environment monitored to determine if the actions were successfully executed.
Figure~\ref{fig:dwr-planstreams} shows a plan produced for the Dock-Worker 
Robots problem depicted in figure~\ref{fig:dwr-problem} with different streams
belonging to the different agents in the domain. 

Note that in the case of joint operators, all the agents involved need a 
``stronger'' synchronisation than what causal links entail. 
For instance in a robot-robot synchronisation they may need to
set some rendezvous points so as to exchange information 
just before starting. This may also involve visual servoing, 
both in robot-robot and human-robot joint operators.

%
%
%
%

\section{HATP in an HRI context}
As highlighted by \cite{bib:alili} one challenge in robotics is to develop socially
interactive and cooperative robots. The meaning of socially interactive
robots is defined in \cite{bib:fong} which states that they must
``operate as partners, peers or assistants, which means 
that they need to exhibit a certain degree of adaptability and flexibility 
to drive the interaction with a wide range of humans''. \cite{bib:klein}
implemented that in what they called ``ten challenges for human 
robot teamwork''. We are convinced that task planners can take care of several
of these challenges. In this case the robot should be able to \cite{bib:klein}: \textit{(1)} 
signal in what tasks it can/wants to participate; \textit{(2)} act in a 
predictable way to ensure human understanding of what it is doing; 
\textit{(3)} publicise its status and its intentions; \textit{(4)} 
negotiate on tasks with its human partner in order to determine roles 
and decide how to perform the tasks; and \textit{(5)} deal with social 
conventions, as well as its human partner's abilities and preferences.
 
To address some of those challenges HATP includes mechanisms to
filter plans so as to keep only those suitable for HRI.
To this end, HATP allows the specification of the following filtering criteria.
\begin{description}
\item[Wasted time:] Avoids plans where an agent(s) mentioned in a plan 
spends a lot of its time waiting.
\item[Effort balancing:] Avoids plans where efforts are not fairly distributed 
among the agents mentioned in a plan.
\item[Control of intricacy:] Avoids plans with too many interdependencies
between the actions of agents mentioned in the plan, as a problem with executing 
just one of those actions could invalidate the entire plan.
\item[Undesirable sequences:] Avoids plans that violate specific user-defined
sequences. 
\end{description}

Combining some of the above criteria could help yield the following 
interesting behaviours: \textit{(1)} the human ends up doing a lot of the 
tasks, but yet the overall effort \cite{bib:alili} taken to do them is significantly lower
than what the robot puts to do a lower number of effort-intensive tasks; 
\textit{(2)} avoiding, when possible, having the human wait for the robot
several times, which essentially prevents the streams from having too
many causal links between them. The filtering criteria are implemented 
by looking through all the plans produced and filtering out the ones that 
do not meet the requirements specified. In the future we intend to study
algorithms that do such filtering online, rather than after primitive 
solutions are found.


\section{Interleaving with geometric reasoning}
\label{sec:interleave}

While an HTN hierarchy allows one to intuitively reason about high-level
tasks such as \code{Transport} in terms of more specific tasks, and eventually 
in terms of basic actions, these still ``abstract out'' the lowest possible 
level of detail by making certain assumptions about the world. For example,
HATP
operator \code{Move} in listing \ref{lst:action} assumes that as long as
location \code{To} is adjacent to location \code{From}, and \code{To} is
not occupied, that the robot at \code{From} will be able to navigate to 
location \code{To}. Clearly, this may not always work for various reasons,
such as there being an obstacle in the path, or certain geometrical 
characteristics of the robot and the connecting path making
the move physically impossible. Combining HATP---and symbolic/task planning in 
general---with the geometric planning algorithms used in robotics
is therefore essential to be able to obtain primitive solutions that 
are viable in the real world.

The work in \cite{silvaIROS13,silvaICRA14,Silva13} 
proposes an interface between HATP and a 
geometric planner. This interface is mainly provided via ``evaluable 
predicates''---predicates in HATP preconditions that are evaluated 
by calling associated external procedures. Such a predicate evaluating to
$true$ amounts to a geometric solution existing for the ``geometric task'' that 
the predicate represents, and evaluating to $false$ amounts to the
non-existence of such a solution. For example, 
the precondition of an HATP action that gives an object
to a person might have an evaluable predicate that invokes the geometric
planner to check the feasibility of the task of giving the object to the
person, and to store the resulting geometric trajectory if any.
This notion of a geometric task is something that is both important in order
to have a meaningful link between the two planning approaches, and also 
specific to the type of geometric planner used. A geometric task essentially
corresponds to one or more motion planning goal-configurations, computed
(automatically) by the geometric planner by taking into account
various criteria such as the visibility and
reachability of objects from the perspectives of different robots
and humans 
in the domain. Aptly called Geometric Task Planner (GTP) \cite{Pandey_2012_GTP},
this planner liberates HATP from having to reason in terms of low-level
details such as grasps and orientations. Using this particular planner
for forming the link with HATP also makes the interface proposed 
different to other interfaces in the literature, such as 
\cite{Dornhege09,Dornhege09icaps,Karlsson12,Lagriffoul12}.

The interface between HATP and the GTP is used to interleave their planning 
algorithms. In one approach, whenever the GTP is invoked by HATP while testing
an evaluable predicate, the non-existence of a GTP solution for the 
associated geometric task (from the current geometric world state) does not 
lead to the 
predicate evaluating to $false$; instead, the GTP backtracks to try 
alternative solutions for the previously invoked geometric tasks 
in an effort to make a solution
possible for the most recently invoked one. Since this may cause changes
to intermediate geometric world states, this approach comes with
mechanisms to ensure that such changes do not affect the 
symbolic world state in a way that invalidates the HATP 
plan being pursued. Such mechanisms are, however, not necessary in the
second approach to interleaved planning that the authors present. 
Here, whenever the GTP cannot
find a solution for a geometric task, it does not---as before---backtrack 
to find alternatives for previous geometric tasks, 
but instead immediately returns with ``failure''. If this leads to
HATP backtracking, HATP
then has the option to try, intuitively, a different ``instance'' 
of the action that needs to be ``undone'' as a consequence of the backtrack
(in addition to the standard option of trying different \textit{actions}); 
this different ``instance'' is basically the same HATP action that needs
to be undone, but this time with a different geometric solution attached
to it.

\begin{figure*}[ht]
\centering
\includegraphics[width=0.7\textwidth]{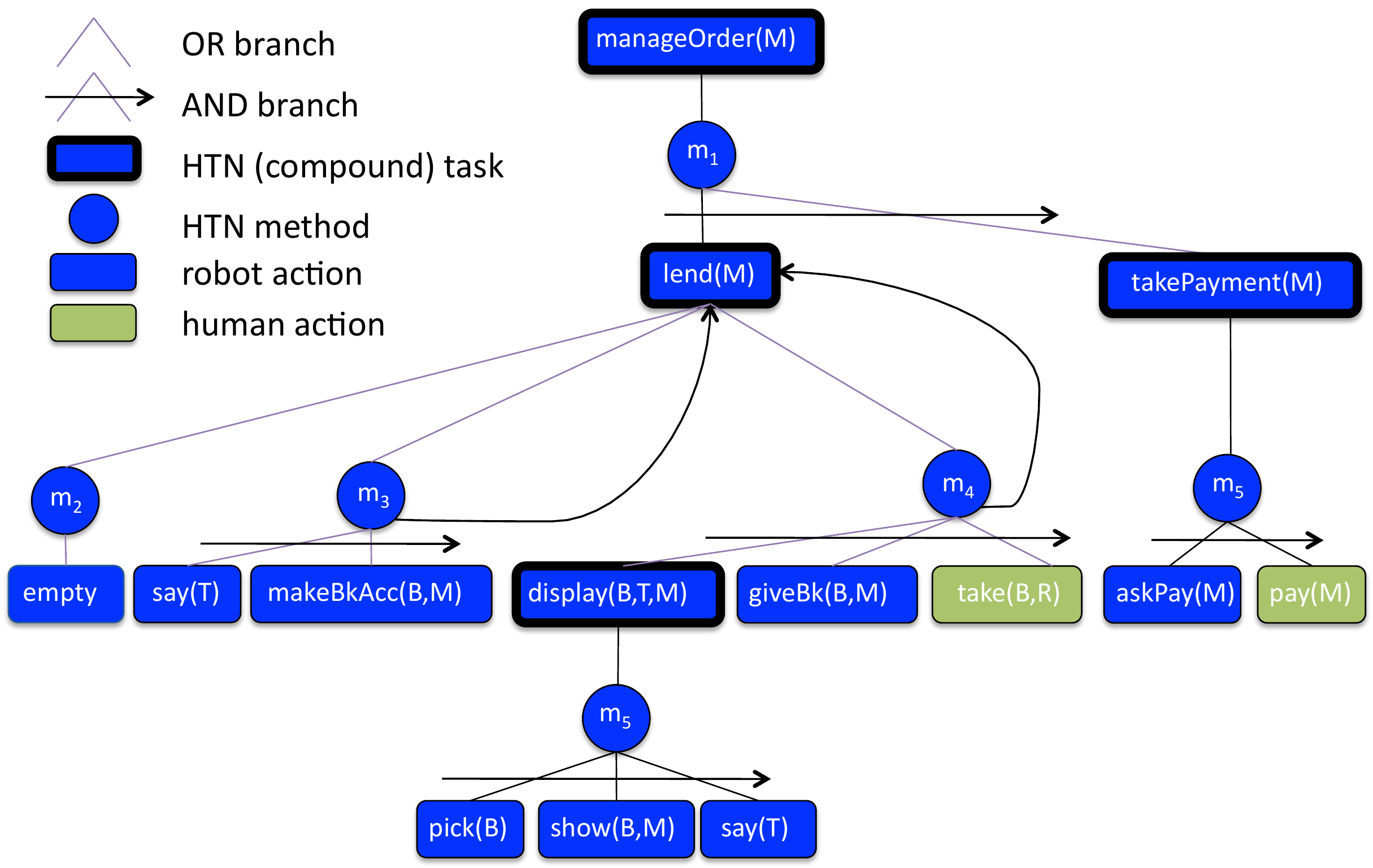}
\caption{This figure depicts a simplified version of the library
domain in \cite{silvaICRA14}. Members ($M$) reserve library books ($B$) online and then come
in person to pick them up from the PR2 ($R$). The PR2 manages the order placed 
for the books by 
recursively lending all the books ordered until there are no more books to 
lend---in which case it will stop the recursion by choosing method $m_2$---and 
then taking payment from the member. Books can either be made accessible
on the table (\code{makeBkAcc}) via method $m_3$, or displayed to the member 
and then given to his/her hand via methods $m_5$ and $m_4$. Action 
\code{say} involves speaking out the title ($T$) of the book.}
\label{fig:hatp-gtp}
\end{figure*}

An interesting feature of the GTP is its ability to plan not just the 
robots' tasks/actions but also the humans', by taking into account their
respective kinematic models. This makes way for the multiple robots/agents
defined in an HATP domain to have a clear association with those defined in 
the GTP domain. For example, figure \ref{fig:hatp-gtp} shows a simplified library
domain \cite{silvaICRA14} where a PR2 robot serves a human customer, consisting of both human and
robot actions. While the PR2-actions will be planned by the GTP from the
perspective of the PR2 (using its kinematic model), those of the
human, which involve paying and taking a book, will be planned 
from the human's perspective.  

\section{The planning and execution architecture}

Both HATP and the GTP are part of the larger LAAS 
robotics architecture \cite{fleury97,bib:belief}. 
This architecture has many components.
It uses the Move3D \cite{Simeon_2001_Move3d} 
motion and manipulation planner for representing the robot's version
of the real world in 3D and for doing
geometric task planning. Through various sensors the robot can also
update its 3D world state in real-time. To this end, a tag-based stereo vision
system is used for object identification and localisation, and a Kinect
(Microsoft) sensor for localising and tracking the human. The execution
controller---the Procedural Reasoning System (PRS)\cite{Ingrand1996}---is 
responsible for invoking HATP when
a task needs to be planned, and also executing the resulting primitive 
solution returned by HATP by invoking various actuators via the interface
provided by Genom \cite{fleury97} to the low-level controllers, which
is also the framework used to wrap them into individual well-defined modules.

In the current architecture, PRS receives goals from the environment,
which it validates by checking for things such as whether the goal has 
already been achieved. If the goal is valid, it is sent as a task to
HATP. If HATP (possibly together with the GTP) 
successfully returns a solution, it is then executed by PRS,
by directly executing the robot's actions and indicating in the right
order to other agents, 
via a dialogue module, what actions they need to execute. To execute an action
directly, PRS sends requests to the relevant Genom modules which may result in
the robot or an arm moving, for example. Indeed, the Genom modules may
actually execute the trajectories found and stored by the GTP if it
was invoked by HATP during the planning process. PRS is also able to
confirm whether the robot's actions and those of the other agents were 
successfully executed, by examining the current state of the (symbolic
and geometric) world.

\section{Conclusion and future work}

We have described in this paper the HATP HTN planner, which has been used 
extensively for practical robotics applications in the LAAS architecture 
\cite{laasArchi} over many years \cite{bib:alili,bib:guitton,bib:warnier,silvaIROS13,silvaICRA14}.
We have focussed on describing how HATP is suited for not just 
HTN planning but also planning in the context of Human-Robot Interaction, 
in a multi-agent setting consisting of multiple humans and robots. 
This was based on two main extensions 
to HATP: the ability to handle user-supplied ``social rules'' that specify what 
is appropriate behaviour for the agents in the domain; and 
interleaving the 
HATP planning algorithm with geometric planning algorithms from the 
robotics community. We have also presented the advantages of the
user-friendly syntax and
semantics of HATP using our own encoding of the Dock Worker Robot
domain described in \cite{bib:automated_planning}. 

There has also been some initial efforts toward extending HATP to support 
separately modelling the beliefs of the different agents in the domain 
\cite{bib:belief}. This allows reasoning about what the different agents 
know, including finding conflicting beliefs, and synchronising beliefs by
planning to notify agents when there are inconsistencies between their beliefs. 
Other interesting work
on HATP that is currently underway is formalising its domain representation
language to show its relation with more traditional representations such 
as that used by the SHOP planner \cite{bib:Nau1999}.
Indeed, this involves developing a mapping from
the syntax and notions of HATP to PDDL-like syntax and notions.

In terms of the link between HATP and the GTP, it would
be interesting to compare the two different combined backtracking strategies.
As the authors in \cite{silvaICRA14} 
have stated, this would require completing the implementation of the
system presented in \cite{silvaIROS13} so that it may be compared empirically 
with the system in \cite{silvaICRA14}. An analytical evaluation would also be
useful to understand in what situations/domains one combined backtracking
approach should 
be favoured over the other. Finally, modifying HATP to interleave planning
with execution to make HATP more ``responsive'' to changes in the environment
would make it even more suitable for real-world robotics applications 
\cite{silvaICRA14}.

\bibliography{bibliography.bib}
\bibliographystyle{aaai}
\end{document}